\documentclass[letterpaper, 10 pt, conference]{ieeeconf}
\IEEEoverridecommandlockouts
\usepackage{amsmath,amssymb,graphicx,booktabs}
\usepackage[nocompress]{cite}
\usepackage{url}
\usepackage{etoolbox}
\graphicspath{{figures/}}
\patchcmd{\abstract}{---}{:}{}{}
\title{\LARGE \bf Surgical Kinematics from Monocular Video with Learned Articulated Motion Constraints}
\author{%
\authorblockN{Mehmet Kerem Turkcan, Soham Samal, Zoran Kostic}
\authorblockA{Columbia University, New York, NY, USA\\
\texttt{mkt2126@columbia.edu}, \texttt{ss7563@columbia.edu}, \texttt{zk2172@columbia.edu}}%
}
\begin{document}
\maketitle
\thispagestyle{empty}
\pagestyle{empty}

\begin{abstract}
Objective assessment of robotic surgery uses instrument kinematics, which must be reconstructed when only video is available. We introduce a kinematic reconstruction network for estimating instrument position, orientation and jaw angle from monocular video. Our visual representation combines global attention pooling of frozen DINOv3 features with local pooling at instrument landmarks from fine-tuned SAM 3.1 masks. Our shared Transformer encoder and temporal convolutional heads integrate this representation with mask geometry, monocular depth and visual state estimates from arm-specific multilayer regression networks. Our position branch predicts displacement magnitude and direction separately to preserve traveled distance. We fit trajectories to predicted state observations and motion increments by differentiable weighted least squares, expressing quaternion observations relative to cumulative predicted rotations to obtain a quadratic orientation objective. We evaluate reconstruction across 2,802 Open-H episodes. Compared with LiveMAE on the main Open-H benchmark, our method reduces path-length mean absolute error from 0.45 to 0.34\,cm and increases temporal mean average precision for motion segmentation from 44.54\% to 54.44\%.
\end{abstract}

\section{Introduction}
Objective assessment of robotic surgery uses path length, speed, smoothness and working volume computed from instrument kinematics~\cite{Ballo2026,Balasubramanian2015}. Reconstructing these kinematics from video enables motion analysis when recorded robot states are unavailable. Surgical recordings are often monocular~\cite{Chen_2025}, and commercial robotic platforms do not routinely provide access to synchronized kinematic data~\cite{10.1007/978-3-031-43996-4_68}.

Appearance changes and occlusion complicate reconstruction. Small position errors can accumulate in path length, while temporal averaging can attenuate motion. Both state accuracy and motion consistency are therefore needed.

\begin{figure*}[t]
\centering
\includegraphics[width=\textwidth]{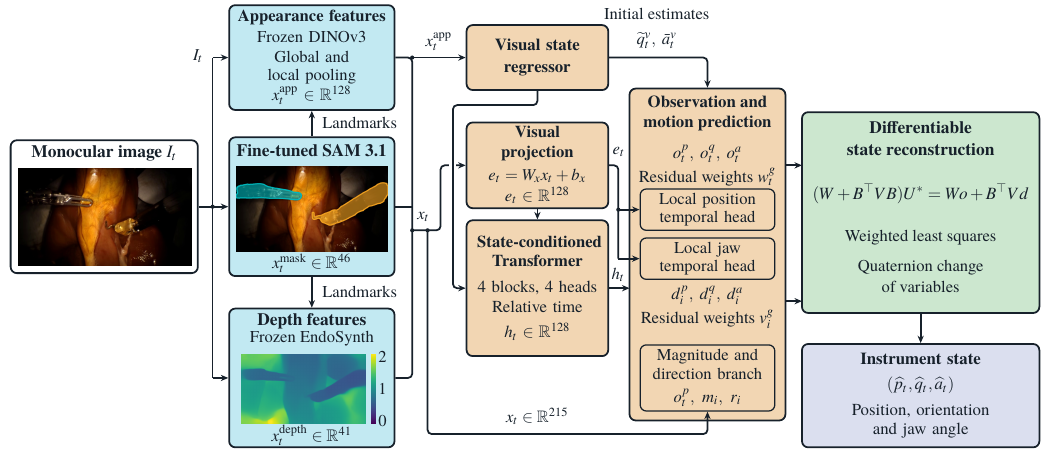}
\caption[Monocular reconstruction of instrument state]{SAM masks define geometry and landmarks for local DINOv3 pooling and depth
sampling. The visual state regressor supplies initial orientation and jaw
estimates to the Transformer and prediction heads. Visual projection maps
$x_t$ to $e_t=W_xx_t+b_x\in\mathbb R^{128}$. Local position and jaw heads
combine $e_t$ with Transformer features $h_t$. The magnitude and direction
branch separately receives $x_t$, corrects position observations and predicts
displacement magnitude $m_i$ and direction correction $r_i$.
Prediction heads produce observations $o_t^g$, increments
$d_i^g$ and residual weights $w_t^g,v_i^g$ for $g\in\{p,q,a\}$.
The displayed normal equation reconstructs position and jaw; orientation follows
Fig.~\ref{fig:constraints}(c). Matrices $W,V$ weight observation and motion
residuals; $B$ forms consecutive differences. Recorded states supervise
predictions; motion constraints stop at gaps. Cyan and yellow identify the
grasper and hook; depth is divided by its image median.
}
\label{fig:pipeline}
\end{figure*}

Our reconstruction network combines a shared Transformer encoder and local
temporal convolutional heads to estimate instrument position, orientation, jaw
angle and motion increments. It integrates appearance, SAM 3.1 mask geometry,
monocular depth and initial visual state estimates (Fig.~\ref{fig:pipeline}).
A position branch predicts displacement magnitude separately from direction,
with supervision on local motion and episode path length.

We fit trajectories by differentiable weighted least squares, using predicted
weights for observation and motion residuals. Cumulative predicted rotations
define a quaternion change of variables, yielding a quadratic orientation
objective with a unique minimizer. We evaluate reconstruction and downstream
motion analysis on Open-H~\cite{consortium2026openhembodimentlargescaledatasetenabling}
using published methods and component ablations.

\section{Related Work}
\textbf{SAM for surgical segmentation.} SurgicalSAM uses prototype prompts~\cite{Yue_2024}, SurgicalPart-SAM adds instrument-region text prompts~\cite{yue2024surgicalpartsamparttowholecollaborativeprompting}, and Surgical SAM 2 prunes temporal memory~\cite{liu2025surgicalsam2realtime}.
Recent work evaluates SAM 3 prompting~\cite{dong2025segmentationbenchmarkingsam3} and adapts it through low-rank updates~\cite{liu2026parameterefficientadaptationsam3promptdriven}.
We fine-tune SAM 3.1 to obtain instrument masks, then extract landmarks for local appearance pooling and trajectory reconstruction.

\textbf{Surgical geometry and state reconstruction.}
Instrument pose estimation uses landmarks~\cite{Du_2018} and geometric models~\cite{Allan_2018}; SurgRIPE~\cite{Xu_2025} benchmarks this task. SurgiPose reconstructs kinematics by matching rendered and observed images~\cite{Chen_2025}; LiveMAE learns joint image and kinematics reconstruction for skill assessment~\cite{10.1007/978-3-031-43996-4_68}. Our trajectory objective constrains both state observations and motion increments.

\textbf{Learned state estimation.}
Kalman filtering~\cite{Kalman_1960} and Rauch, Tung and Striebel (RTS) smoothing~\cite{RAUCH_1965} couple observations through a dynamical model; associative prefix operations enable parallel filtering and smoothing~\cite{Sarkka_2021}. DROID-SLAM integrates learned estimation with differentiable bundle adjustment~\cite{NEURIPS2021_89fcd07f}; Theseus provides general differentiable least squares~\cite{NEURIPS2022_18596929}. We construct quadratic trajectory objectives, using a quaternion change of variables to retain this structure for orientation.

\section{Method}
\label{sec:method}
\subsection{Video-to-state reconstruction}
Given images $I_t$ and timestamps $\tau_t$, we estimate the state of each patient-side manipulator, or PSM, in its recorded base frame:
\begin{equation}
s_t=(p_t,q_t,a_t),\qquad p_t\in\mathbb R^3,\quad q_t\in\mathbb S^3,\quad a_t\in\mathbb R.
\end{equation}
Position is measured in meters and jaw angle in radians; $q_t$ is a unit quaternion. Our temporal reconstruction network estimates each arm's kinematics independently with shared parameters. Recorded instrument positions, orientations and jaw angles provide training targets and evaluation references.

\subsection{Visual representation}
\textbf{Instrument segmentation and landmarks.}
We fine-tune SAM 3 and SAM 3.1~\cite{ICLR2026_e0982cbc} for surgical segmentation using
images from CholecInstanceSeg, CholecSeg8k, DSAD and
Endoscapes~\cite{Alabi2025,hong2020cholecseg8ksemanticsegmentationdataset,Carstens2023,murali2024endoscapesdatasetsurgicalscene}.
For SAM 3.1, we expand the fine-tuning data with CRCD anatomy annotations~\cite{Oh_2025}
and uTenn labels for the instrument body, wrist and left and right gripper regions.
Segmentation training and kinematic
evaluation use separate source groups. We select the segmenter by validation
mask average precision (AP) and freeze it before kinematic reconstruction.

We assign instrument masks to each PSM by image entry point and track each
instrument through entry-point and centroid continuity.
The vector $x_t^{\rm mask}\in\mathbb R^{46}$ encodes instrument extent, axis,
region geometry, segmentation scores and mask availability.

\begin{figure*}[t]
\centering
\includegraphics[width=\textwidth]{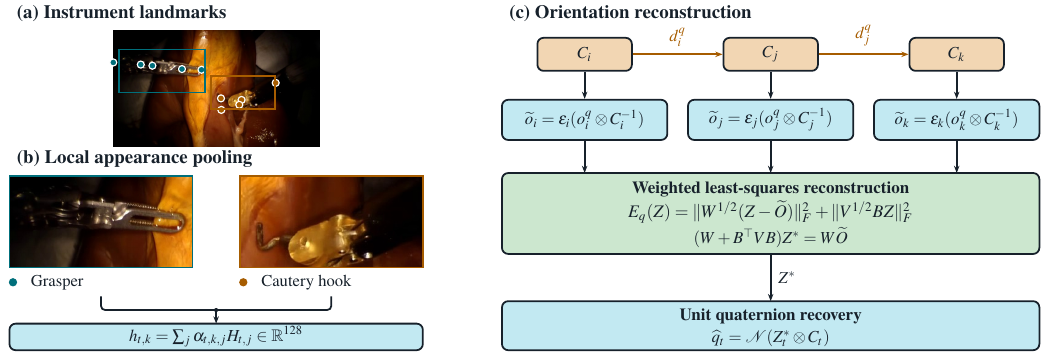}
\caption{(a,b) SAM-derived landmarks define local appearance descriptors $h_{t,k}$.
(c) Rotation increments accumulate as $C_j=C_i\otimes d_i^q$ from the identity
within each recording interval. Signs $\epsilon_t\in\{-1,1\}$ align transformed
observations to the first in the interval. With stacked observations
$\widetilde O$, diagonal observation and motion weights
$W_{tt}=4w_t^q/s_q^2$, $V_{ii}=4\lambda_qv_i^q/s_{\Delta q}^2$ and
within-interval difference operator $B$, four tridiagonal linear systems yield
$Z^*$. The inverse change of variables and normalization recover unit quaternions.}
\label{fig:constraints}
\end{figure*}

\textbf{Global and local appearance features.}
Our visual feature network represents instruments and surrounding tissue
using frozen DINOv3 ViT-L/16 features~\cite{siméoni2025dinov3}.
A fixed Gaussian projection to 128 dimensions and two 128-channel $3\times3$
convolutions produce spatial vectors
$H_{t,j}\in\mathbb R^{128}$ at normalized image locations
$u_j^{\rm img}\in[0,1]^2$.
A $1\times1$ convolution predicts arm-specific spatial attention, which pools
these vectors into the global descriptor
$g_t\in\mathbb R^{128}$. SAM masks define six landmarks:
the instrument centroid, distal and entry endpoints of its principal axis,
and the centroids of the wrist and two gripper regions.
At landmark $\ell_{t,k}$, we pool nearby image features with Gaussian weights:
\begin{equation}
 \begin{aligned}
 h_{t,k}&=\sum_j\alpha_{t,k,j}H_{t,j},\\
 \alpha_{t,k,j}&=\operatorname{softmax}_j
 \left(-\frac{\|u_j^{\rm img}-\operatorname{clip}(\ell_{t,k},0,1)\|_2^2}{2\sigma^2}\right).
 \end{aligned}
\end{equation}
The scale $\sigma$ controls pooling extent. A shared fusion network
$f_{\rm fuse}$ with a 256-unit hidden layer combines global and local features:
\begin{equation}
 x_t^{\rm app}=g_t+f_{\rm fuse}(g_t,h_{t,1},\ldots,h_{t,6},c_t,b_t)
 \in\mathbb R^{128}.
\end{equation}
Here $c_t$ contains image-centered landmark coordinates and $b_t$ marks
available landmarks. Missing landmarks use $h_{t,k}=g_t$ and zero coordinates.
Arm-specific multilayer regression networks with 128-unit hidden layers predict
position, orientation and jaw angle from this descriptor. We center position
targets within training episodes and jointly train the convolutional layers,
fusion network and regressors against recorded states. We then freeze these
networks; their orientation and jaw estimates initialize and condition temporal
refinement.
Fig.~\ref{fig:constraints}(a,b) illustrates landmarks and local pooling.

\textbf{Monocular depth.}
We estimate relative depth using the frozen EndoSynth Depth Anything V1
ViT-B/14 network~\cite{10.1007/978-3-032-08009-7_12}, fine-tuned on synthetic
surgical images and depth maps. At each landmark, we take the median depth of
nearby pixels within the instrument mask, when available. The descriptor
$x_t^{\rm depth}\in\mathbb R^{41}$ combines landmark depths, ratios to median
image depth, interlandmark differences, instrument and scene statistics,
products of depth and image-centered coordinates, and availability indicators.
Sec.~\ref{sec:experiments} evaluates further decoder fine-tuning on rendered Open-H.
Together, appearance, mask geometry and depth form the temporal input:
\begin{equation}
 x_t=[(x_t^{\rm app})^\top,(x_t^{\rm mask})^\top,
      (x_t^{\rm depth})^\top]^\top\in\mathbb R^{215}.
\end{equation}
We impute missing entries and normalize features using training statistics. External-camera experiments
use global appearance with the same temporal formulation.
\subsection{Learned articulated motion constraints}
\textbf{State and motion prediction.}
Our shared Transformer encoder and prediction heads estimate instrument position,
orientation and jaw observations and increments. Position and jaw angle are
standardized. The frozen visual regressor supplies jaw estimate $\bar a_t^v$ and
quaternion $q_t^v$, whose sign we align to a training-derived reference
$q_{\rm anchor}$, giving $\widetilde q_t^v$.
The encoder combines projected visual features
$e_t=W_xx_t+b_x\in\mathbb R^{128}$ with initial orientation, jaw and relative
frame time. Four blocks with four attention heads, feature width 128 and
feedforward width 512 produce $h_t\in\mathbb R^{128}$:
\begin{align}
 u_t&=e_t+W_\tau\phi(\tau_t-\tau_0)
       +W_s[\widetilde q_t^v;\bar a_t^v],\nonumber\\
 h_{1:T}&=\mathcal T(u_{1:T}).
 \label{eq:observation_context}
\end{align}
Here $\phi(\tau_t-\tau_0)$ sinusoidally encodes time relative to the first frame.
Attention spans all retained frames. A shared observation head with a 128-unit
hidden layer predicts position and refines visual jaw and orientation estimates:
\begin{align}
 o_t^{p,0}&=f_p(h_t),\qquad o_t^{a,0}=\bar a_t^v+f_a(h_t),\nonumber\\
 o_t^q&=\mathcal N\bigl(\widetilde q_t^v+f_q(h_t)\bigr),
 \qquad \mathcal N(z)=z/\|z\|_2.
 \label{eq:observations}
\end{align}
A linear head predicts observation weights. Our motion head has a 128-unit
hidden layer and receives $\eta_{ij}=[h_i;h_j;h_j-h_i;\theta_{ij}]$, where
$\theta_{ij}$ encodes elapsed time $\tau_j-\tau_i$. It predicts position and jaw
increments $d_i^{p,0},d_i^{a,0}$, rotation vectors and motion weights.
Rotation vectors yield unit-quaternion increments $d_i^q$ with target
$q_i^{-1}\otimes q_j$; $\otimes$ denotes quaternion multiplication.
The set $\mathcal A$ contains consecutive frame pairs within recording intervals.

\textbf{Position and jaw refinement.}
Our temporal convolutional heads refine position from $[e_t;h_t]$ and jaw from
$[e_t;h_t;\bar a_t^v;o_t^{a,0}]$. Both use 64 channels and residual blocks with
layer normalization, size-three depthwise temporal convolution and channel mixing.
Dilations are one and two for position; one, two, four and eight for jaw.

The resulting local features $\zeta_t^g\in\mathbb R^{64}$ feed a linear
observation head and a motion head with a 64-unit hidden layer. For
$g\in\{p,a\}$, their residual corrections give
\begin{align}
 \bar o_t^g&=o_t^{g,0}+r_g(\zeta_t^g),\nonumber\\
 \bar d_i^g&=d_i^{g,0}+s_{\Delta g}\odot
 r_{\Delta g}([\zeta_i^g;\zeta_j^g;\zeta_j^g-\zeta_i^g;\theta_{ij}]).
 \label{eq:position_refinement}
\end{align}
The scale $s_{\Delta g}$ contains training root-mean-square increments in
standardized coordinates; $\odot$ denotes coordinatewise multiplication.
These heads also refine the scalar outputs used to weight observation and motion residuals.
Temporal convolutions and motion constraints stop at recording gaps.
For jaw reconstruction, $o_t^a=\bar o_t^a$ and $d_i^a=\bar d_i^a$.

\textbf{Displacement magnitude and direction.}
To constrain traveled distance directly, our position branch estimates
displacement magnitude separately from direction. It projects $x_t$ to 96 channels
and applies four residual temporal convolutional blocks with dilations one, two,
four and eight. A linear head adds a standardized position correction to
$\bar o_t^p$, yielding $o_t^p$. A motion head with
a 96-unit hidden layer uses the same pairwise feature construction as the shared
head to predict a direction correction $r_i$ and positive magnitude $m_i$ in meters.

Let $S_p$ contain training position standard deviations and $G_p$ positive
validation-calibrated gains, both diagonal. The initial displacement in meters
is $\widetilde d_i^p=S_pG_p\bar d_i^p$. We normalize its corrected direction and
convert the predicted displacement to standardized coordinates:
\begin{equation}
 d_i^p=S_p^{-1}m_i\mathcal N(\widetilde d_i^p+r_i).
 \label{eq:magnitude}
\end{equation}
A softplus output scaled by the training root-mean-square displacement gives
$m_i$. Corrected observations $o_t^p$ and increments $d_i^p$ enter weighted
reconstruction.

\textbf{Weighted least-squares reconstruction.}
Positive coefficients $w_t^g,v_i^g$ weight observation and motion residuals.
We bound them and divide both by the episode mean observation coefficient for
each group $g\in\{p,q,a\}$. Position and jaw trajectories minimize
\begin{align}
 E_g(U)={}&\sum_t w_t^g\|U_t-o_t^g\|_2^2\nonumber\\
 &+\lambda_g\sum_{(i,j)\in\mathcal A}\sum_k
 \frac{v_i^g}{s_{\Delta g,k}^2}
 (U_{j,k}-U_{i,k}-d_{i,k}^g)^2,
 \label{eq:euclidean}
\end{align}
where $g\in\{p,a\}$ and $s_{\Delta g,k}>0$ is coordinate $k$ of the increment
scale. The positive coefficient $\lambda_g$ balances observation and motion;
we learn $\lambda_p$ with the position head and keep $\lambda_a,\lambda_q$ fixed.
For each coordinate, define $(BU)_{(i,j)}=U_j-U_i$ and diagonal weights
$W_{tt}=w_t^g$, $V_{ii}=\lambda_gv_i^g/s_{\Delta g,k}^2$.
The minimizing trajectory satisfies
\begin{equation}
 (W+B^\top VB)U^*=Wo+B^\top Vd.
 \label{eq:normal}
\end{equation}
For orientation, cumulative rotations $C_j=C_i\otimes d_i^q$ start from
$C_{t_0}=1$ within each continuous interval. The change of variables
$\widetilde o_t=\epsilon_t(o_t^q\otimes C_t^{-1})$ removes these rotations;
$\epsilon_t\in\{-1,1\}$ aligns signs to the first transformed observation.
The transformed trajectory $Z^*\in\mathbb R^{T\times4}$ minimizes
\begin{align}
 E_q(Z)={}&
 \sum_t\frac{4w_t^q}{s_q^2}\|Z_t-\widetilde o_t\|_2^2\nonumber\\
 &+\lambda_q\sum_{(i,j)\in\mathcal A}
 \frac{4v_i^q}{s_{\Delta q}^2}\|Z_j-Z_i\|_2^2.
 \label{eq:quaternion}
\end{align}
The positive scales $s_q^2$ and $s_{\Delta q}^2$ derive from training means
of $\ell_q(q_t,q_{\rm anchor})$ and
$\ell_q(q_i,q_j)$, where $\ell_q(u,v)=4(1-(u^\top v)^2)$.

\textbf{Proposition 1.}
With positive observation coefficients and nonnegative motion coefficients,
Eqs.~\ref{eq:euclidean} and~\ref{eq:quaternion} have unique minimizers
obtained by solving independent tridiagonal linear systems.

\textit{Proof.}
For $Q_t=Z_t\otimes C_t$, right multiplication by a unit quaternion gives
$\|Q_j-Q_i\otimes d_i^q\|_2=\|Z_j-Z_i\|_2$.
For either objective, its scaled weights give
$\xi^\top(W+B^\top VB)\xi
=\|W^{1/2}\xi\|_2^2+\|V^{1/2}B\xi\|_2^2>0$ for nonzero $\xi$.
Since $\mathcal A$ connects consecutive frames, the linear system for each
coordinate is tridiagonal.

From the unconstrained minimizer, we recover unit quaternions by applying the
inverse change of variables and normalizing:
$\widehat q_t=\mathcal N(Z_t^*\otimes C_t)\in\mathbb S^3$.
We solve these systems in parallel~\cite{Zhang_2010} and differentiate through
reconstruction. At inference, we apply a positive affine calibration to reconstructed jaw
angles, then project them onto the interval spanned by training jaw angles across
both arms. We rescale position and jaw angle to physical units.

\textbf{Training objectives.}
We supervise observations and reconstructed states with normalized squared position,
sign-invariant quaternion and jaw Huber losses. We gradually introduce motion losses
on predicted increments, reconstructed position increments and multiscale path
lengths, including complete episodes, without crossing gaps. Translation calibration
subtracts the mean first-second position residual. Pointwise and motion losses
use all retained frames; aligned position supervision and state evaluation use
later frames (Sec.~\ref{sec:evaluation}).

With the Transformer encoder and initial prediction heads frozen, we fit position
and jaw refinement heads using multiscale endpoint and path-length losses for position, and absolute observation
and reconstructed-angle errors for jaw. Validation selects initial position
gains and, on main endoscopy, affine jaw calibration.

Finally, we freeze these components and fit the displacement branch and
$\lambda_p$. Losses supervise displacement magnitude, multiscale endpoint
displacements in observations and reconstructed positions, aligned position error,
and absolute error in the mean episode path length of both arms. Orientation and jaw
predictions remain fixed.

\section{Data and Evaluation}
\label{sec:evaluation}
\textbf{Open-H evaluation cohorts.}
Table~\ref{tab:data} summarizes paired video and robot states from
Open-H~\cite{consortium2026openhembodimentlargescaledatasetenabling}.
Episodes are released recording windows. Main endoscopy from Johns Hopkins
University, or JHU, covers cholecystectomy training, porcine cholecystectomy and
cautery. After tracking review, we retain intervals with landmark coverage,
visible motion and agreement with recorded states, using identical timestamps
for both arms. We include every released episode of JHU central airway
obstruction, or CAO, resection and Hamlyn manipulation and suturing.

\begin{table}[!ht]
\centering\small
\renewcommand{\arraystretch}{1.12}
\caption{Open-H kinematic evaluation. Each frame is paired with states for both arms.
Additional JHU uses main-benchmark networks without further fitting.}
\label{tab:data}
\begin{tabular*}{\columnwidth}{@{\extracolsep{\fill}}lrr@{}}
\toprule
Cohort & Episodes & Frames \\
\midrule
Main endoscopy & 930 & 124,839 \\
CAO & 783 & 52,748 \\
Hamlyn & 1,019 & 552,753 \\
Additional JHU & 70 & 15,661 \\
\midrule
Total & 2,802 & 746,001 \\
\bottomrule
\end{tabular*}
\end{table}

\begin{table*}[t]
\centering\small
\renewcommand{\arraystretch}{1.08}
\caption{Perception and kinematic reconstruction. AP scores fully annotated categories; foreground metrics use complete instrument masks. Depth MAE uses the rendered Open-H test set. Reconstruction uses the main Open-H benchmark. Bold marks the best value per column and panel.}
\label{tab:main}
\begin{minipage}[t]{0.40\textwidth}\centering
\textbf{Mask and box AP}\par\smallskip
\begin{tabular}{@{}lrr@{}}
\toprule
Segmenter & Mask AP $\uparrow$ & Box AP $\uparrow$ \\
\midrule
Base SAM 3.1 & 2.77 & 3.29 \\
Fine-tuned SAM 3.1 & \textbf{37.07} & \textbf{41.44} \\
\bottomrule
\end{tabular}
\end{minipage}\hfill
\begin{minipage}[t]{0.34\textwidth}\centering
\textbf{Instrument foreground}\par\smallskip
\begin{tabular}{@{}lrr@{}}
\toprule
Segmenter & IoU $\uparrow$ & Dice $\uparrow$ \\
\midrule
SurgicalSAM~\cite{Yue_2024} & 73.85 & 82.23 \\
Fine-tuned SAM 3.1 & \textbf{85.57} & \textbf{90.98} \\
\bottomrule
\end{tabular}
\end{minipage}\hfill
\begin{minipage}[t]{0.23\textwidth}\centering
\textbf{Rendered depth}\par\smallskip
\begin{tabular}{@{}lr@{}}
\toprule
Network & MAE $\downarrow$ \\
\midrule
EndoSynth & 0.9114 \\
Fine-tuned & \textbf{0.2454} \\
\bottomrule
\end{tabular}
\end{minipage}
\par\medskip\textbf{Instrument-state reconstruction}\par\smallskip
\begin{tabular*}{\textwidth}{@{\extracolsep{\fill}}llrrrrr@{}}
\toprule
Method & Input & Rigid $\downarrow$ & Calibrated $\downarrow$ & Orientation $\downarrow$ & Jaw $\downarrow$ & Path $\downarrow$ \\
 & & cm & cm & deg & deg & cm \\
\midrule
MS-TCN~\cite{Farha_2019_CVPR} & Global DINOv3 & 0.0905 & 0.2086 & 13.00 & 3.08 & 0.7275 \\
PatchTST~\cite{Yuqietal-2023-PatchTST} & Global DINOv3 & 0.0665 & 0.1546 & 15.68 & 3.92 & 0.4815 \\
Transformer encoder, 4 blocks~\cite{NIPS2017_3f5ee243} & Global DINOv3 & 0.0625 & 0.1457 & 14.31 & 3.85 & 0.4749 \\
LiveMAE~\cite{10.1007/978-3-031-43996-4_68} & RGB & 0.0638 & 0.1536 & 12.59 & 3.25 & 0.4510 \\
\textbf{Proposed} & SAM, DINOv3, depth & \textbf{0.0528} & \textbf{0.1209} & \textbf{8.78} & \textbf{2.31} & \textbf{0.3392} \\
\bottomrule
\end{tabular*}
\end{table*}

\begin{figure*}[t]
\centering
\includegraphics[width=\textwidth]{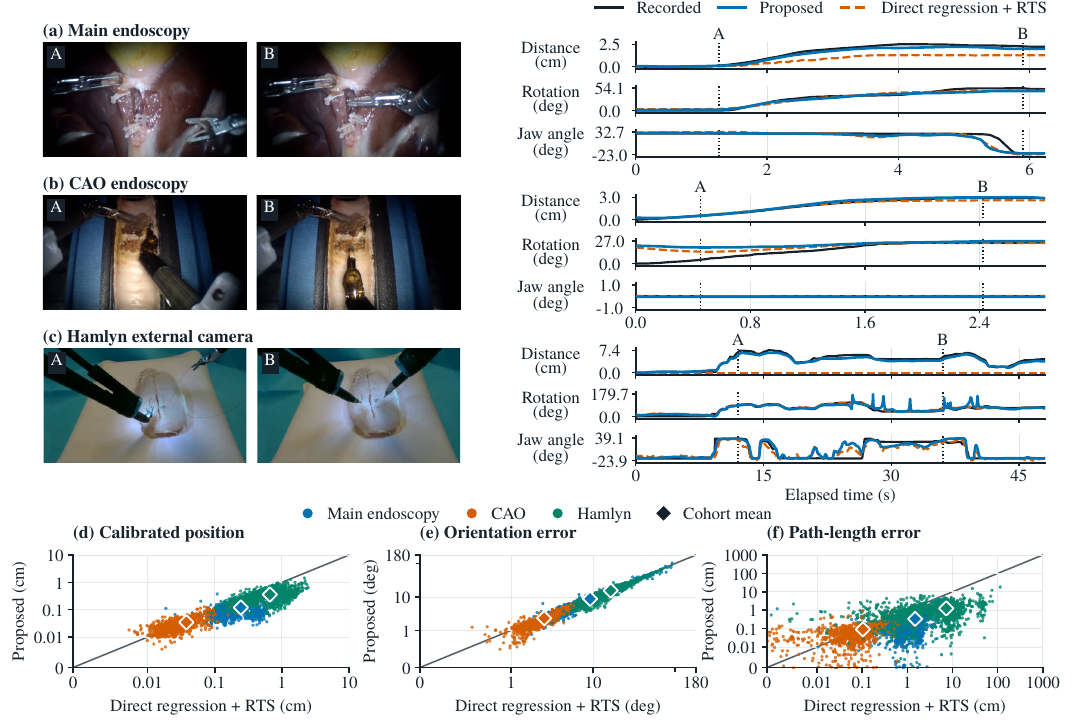}
\caption[Instrument trajectories and state accuracy]{(a,b,c) PSM1 examples selected using recorded motion and visible activity.
Position uses first-second translation calibration; distance and rotation are
displacements from the recorded initial state. Guides identify frames A and B.
(d,e,f) Paired episode errors across the three Open-H cohorts. Direct regression
+ RTS uses the same visual inputs and omits weighted reconstruction and position
and jaw refinement.
Points represent episodes; diamonds mark cohort means. Values below the diagonal
favor our method. Axes are linear through one degree or 0.01\,cm, then logarithmic.}
\label{fig:trajectories}
\end{figure*}

\begin{table}[t]
\centering\small
\renewcommand{\arraystretch}{1.08}
\setlength{\tabcolsep}{1pt}
\caption{Temporal estimation and component contributions. Regressors use our visual inputs, direct state supervision and selected RTS. Training ablations precede position and jaw refinement; component removals use the complete checkpoint without refitting. Populations follow Table~\ref{tab:main}; bold marks the lowest error.}
\label{tab:ablation}
\begin{tabular*}{\columnwidth}{@{\extracolsep{\fill}}lrrrrr@{}}
\toprule
Method & Rigid $\downarrow$ & Calib. $\downarrow$ & Orient. $\downarrow$ & Jaw $\downarrow$ & Path $\downarrow$ \\
 & cm & cm & deg & deg & cm \\
\midrule
\multicolumn{6}{@{}l}{\textit{Alternative temporal regressors}} \\
GRU, 2 layers~\cite{cho-etal-2014-learning} & 0.0569 & 0.1357 & 8.98 & 2.52 & 0.4069 \\
LSTM, 2 layers~\cite{Hochreiter_1997} & 0.0641 & 0.1499 & 9.07 & 2.49 & 0.4280 \\
Dilated TCN~\cite{bai2018empiricalevaluationgenericconvolutional} & 0.0632 & 0.1422 & 8.80 & 2.48 & 0.4090 \\
MS-TCN~\cite{Farha_2019_CVPR} & 0.0632 & 0.1438 & 8.89 & 2.45 & 0.4070 \\
PatchTST~\cite{Yuqietal-2023-PatchTST} & 0.0581 & 0.1285 & 9.05 & 2.56 & 0.4078 \\
Transformer encoder~\cite{NIPS2017_3f5ee243} & 0.0563 & 0.1261 & 9.00 & 2.50 & 0.3692 \\
\midrule
\multicolumn{6}{@{}l}{\textit{Training before position and jaw refinement}} \\
Direct state regression & 0.1082 & 0.2446 & 9.89 & 2.73 & 1.4836 \\
Direct regression + RTS & 0.1082 & 0.2446 & 9.21 & 2.62 & 1.4836 \\
Uniform weights & 0.0534 & 0.1238 & 8.86 & 2.53 & 0.4099 \\
\midrule
\multicolumn{6}{@{}l}{\textit{Component removal at inference}} \\
Neither refinement & \textbf{0.0527} & 0.1219 & \textbf{8.78} & 2.33 & 0.3854 \\
No position refinement & \textbf{0.0527} & 0.1219 & \textbf{8.78} & \textbf{2.31} & 0.3854 \\
No jaw refinement & 0.0528 & \textbf{0.1209} & \textbf{8.78} & 2.33 & \textbf{0.3392} \\
Without reconstruction & 0.0748 & 0.1363 & 9.80 & 2.46 & 10.3219 \\
\textbf{Proposed} & 0.0528 & \textbf{0.1209} & \textbf{8.78} & \textbf{2.31} & \textbf{0.3392} \\
\bottomrule
\end{tabular*}
\end{table}

\textbf{Training and evaluation splits.}
Main endoscopy, CAO and Hamlyn use three-fold cross-validation.
Each run reserves one fold for testing. Every fifth training group forms the
validation set used to select training durations for the Transformer
encoder, prediction heads and subsequent refinement heads. We then refit each
stage for its selected duration on both training folds, using all training episodes.
We select the
displacement branch architecture and training duration on main-cohort validation
data and keep both fixed across all three cohorts.

Each episode's frames and both arms remain together. Main endoscopy and Hamlyn
folds balance settings and tasks, respectively; CAO groups blocks of 20 consecutive
source episodes. Validation preserves this grouping and Hamlyn task balance.
Learned visual features and normalization follow the same training partitions.

\textbf{Additional recordings.}
We evaluate 70 further JHU episodes using fixed reconstruction networks fitted on the first
main-benchmark training split of 620 episodes. These recordings are excluded
from kinematic fitting and checkpoint selection. They retain released video-state pairings.

\textbf{Rendered depth data.}
Open-H Sanoscience renderings are partitioned into 18 training, 4 validation
and 4 test source groups. Each contains six camera views under three
lighting conditions.

\textbf{Video-state synchronization.}
For main endoscopy, a single offset synchronizes each clip using detected instrument
landmarks and recorded kinematics. Cross-validated landmark prediction with
regularized linear regression selects the offset; we retain the released pairing
when no estimate is available. All reconstruction methods use identical reference pairings.
CAO and Hamlyn retain released pairings and timestamps.

\textbf{Position accuracy.}
The first second, $\tau_t<\tau_0+1$\,s, defines calibration interval $\mathcal C$; later frames define scoring interval $\mathcal S$. We align predicted positions to the recorded trajectory by fitting rotation and translation on $\mathcal S$ using Kabsch alignment~\cite{Kabsch_1976}, without rescaling. For translation calibration, we subtract the mean prediction residual on $\mathcal C$. Both transforms are fitted per episode and arm. For transformed prediction $\widetilde p_t$, coordinate mean absolute error (MAE) in centimeters is
\begin{equation}
E_p=\frac{100}{3|\mathcal S|}\sum_{t\in\mathcal S}\|\widetilde p_t-p_t\|_1.
\end{equation}
State metrics require nonempty calibration and scoring intervals.

\textbf{Orientation and jaw accuracy.}
Orientation error in degrees is $(360/\pi)\arccos(|\langle\hat q_t,q_t\rangle|)$ for normalized quaternions. Jaw MAE is also converted to degrees. Both use the same scoring interval. Orientation is evaluated in the recorded PSM base frame, independently of position alignment. State errors are averaged over arms and then episodes.

\textbf{Path length and motion indicators.}
For consecutive frame pairs $\mathcal A$ within continuous recording intervals, path length is
\begin{equation}
L(P)=\sum_{(i,j)\in\mathcal A}\|p_j-p_i\|_2.
\end{equation}
This measurement is invariant to rigid transforms and includes the first second. Prediction and reference use identical timestamps; scoring adds no further smoothing. We average both arm path lengths before computing each episode's absolute error and convert the result to centimeters. We also compare episode
mean speed, log dimensionless jerk and jaw closure count using the same intervals
and arm aggregation~\cite{Ballo2026,Balasubramanian2015}. Log dimensionless jerk
(LDJ) is $-\log(D^5J/L^2)$ for $L,J>0$, where $D$ and $J$ are duration and
integrated squared jerk over continuous intervals of at least eight frames.
LDJ errors use the same reference-defined episodes for all methods.
Jaw closures are downward crossings
of a training-derived jaw threshold.

\textbf{Motion segmentation.}
A two-stage MS-TCN with residual logit refinement~\cite{Farha_2019_CVPR}
segments reconstructed states and derived motion features into idle, open-jaw
motion, jaw closing, closed-jaw motion and jaw opening. The reference classes are
defined from recorded kinematics. For each fold, one classifier trained on recorded
training kinematics evaluates every reconstruction with identical preprocessing,
normalization and decoding.

\section{Experiments}
\label{sec:experiments}
We evaluate how accurately our reconstruction network estimates instrument kinematics
from monocular video and preserves motion for downstream analysis.
The experiments examine (1) surgical segmentation and depth fine-tuning,
(2) reconstruction accuracy, (3) the contribution of reconstruction
and position and jaw refinement, (4) additional Open-H tasks,
camera views and recordings, and (5) objective performance indicators and motion
segmentation from reconstructed kinematics.

\textbf{Surgical perception.}
We compare pretrained SAM 3.1 with the fine-tuned SAM 3.1 segmenter used by our
reconstruction network on identical test images. COCO-style AP scores categories
where annotations are complete; CholecSeg8k evaluation uses the released dense
masks and class mapping. Table~\ref{tab:main} shows that fine-tuning increases
mask AP from 2.77 to 37.07 and box AP from 3.29 to 41.44.

\textbf{Comparison with a surgical segmenter.}
We train SurgicalSAM's prototype prompt encoder and mask decoder with frozen
SAM ViT-H features~\cite{Yue_2024} on the same partitions and complete instrument
annotations; SAM 3.1 additionally learns anatomy, wrist and gripper labels.
We select SurgicalSAM by validation foreground IoU and compare its merged
instrument mask with SAM 3.1 predictions from the \emph{surgical instrument}
prompt. Table~\ref{tab:main} reports mean image IoU and Dice on fully annotated
test images at native resolution; both equal one for empty prediction and target.
Our fine-tuned SAM 3.1 segmenter achieves 85.57\% instrument foreground IoU,
compared with 73.85\% for the SurgicalSAM. We derive reconstruction
landmarks from the predicted instrument masks.

\textbf{Depth fine-tuning on rendered Open-H.}
We fine-tune the EndoSynth depth decoder on rendered Sanoscience images with its
visual encoder frozen, using normalized inverse-depth and spatial-gradient errors.
Table~\ref{tab:main} reports mean scale-and-shift-invariant absolute error on
72 test episodes. Fine-tuning reduces error from 0.9114 to 0.2454, a 73.08\%
reduction, with improvement on every test episode.

\begin{table}[t]
\centering\small
\renewcommand{\arraystretch}{1.08}
\setlength{\tabcolsep}{1.5pt}
\caption{Open-H cohort evaluation. CAO and Hamlyn use separately trained networks; additional JHU retains main-benchmark weights. MS-TCN and Transformer regression use our visual representation; LiveMAE uses RGB. Direct regression shares our inputs and omits weighted reconstruction and position and jaw refinement; RTS smooths the same predictions. Bold marks each cohort's lowest error.}
\label{tab:extensions}
\begin{tabular*}{\columnwidth}{@{\extracolsep{\fill}}lrrrrr@{}}
\toprule
Method & Rigid $\downarrow$ & Calib. $\downarrow$ & Orient. $\downarrow$ & Jaw $\downarrow$ & Path $\downarrow$ \\
 & cm & cm & deg & deg & cm \\
\midrule
\multicolumn{6}{@{}l}{\textit{CAO endoscopy, 783 episodes}} \\
Direct state regression & 0.0142 & 0.0379 & 2.70 & 0.26 & 0.1281 \\
Direct regression + RTS & 0.0139 & 0.0377 & 2.54 & 0.26 & 0.1039 \\
\textbf{Proposed} & \textbf{0.0114} & \textbf{0.0347} & \textbf{2.31} & \textbf{0.20} & \textbf{0.0921} \\
\addlinespace[3pt]
\multicolumn{6}{@{}l}{\textit{Hamlyn external camera, 1,019 episodes}} \\
Direct state regression & 0.4093 & 0.6611 & 17.84 & 7.94 & 7.2639 \\
Direct regression + RTS & 0.4093 & 0.6611 & 16.42 & 7.78 & 7.2639 \\
\textbf{Proposed} & \textbf{0.1902} & \textbf{0.3591} & \textbf{15.55} & \textbf{7.10} & \textbf{1.2627} \\
\addlinespace[3pt]
\multicolumn{6}{@{}l}{\textit{Additional JHU recordings, 70 episodes}} \\
MS-TCN~\cite{Farha_2019_CVPR} & 0.1409 & 0.3079 & 24.91 & 6.62 & 1.1395 \\
LiveMAE~\cite{10.1007/978-3-031-43996-4_68} & 0.1188 & 0.2527 & 30.11 & 6.85 & 0.9684 \\
Transformer encoder~\cite{NIPS2017_3f5ee243} & 0.1086 & 0.2165 & 25.02 & 6.73 & 1.0477 \\
Direct state regression & 0.1458 & 0.2415 & 25.85 & 6.81 & 3.0887 \\
Direct regression + RTS & 0.1458 & 0.2415 & 25.15 & 6.72 & 3.0887 \\
\textbf{Proposed} & \textbf{0.0987} & \textbf{0.2112} & \textbf{24.10} & \textbf{6.49} & \textbf{0.7751} \\
\bottomrule   
\end{tabular*}
\end{table}

\begin{table}[t]
\centering\small
\renewcommand{\arraystretch}{1.08}
\setlength{\tabcolsep}{1.4pt}
\caption{Downstream analysis on the main Open-H benchmark. MS-TCN, PatchTST and Transformer regression use global DINOv3 features; LiveMAE uses RGB images. LDJ denotes log dimensionless jerk. All reconstructions use the same MS-TCN classifier. Temporal mAP averages IoU thresholds $\{0.1,0.2,0.3,0.4,0.5\}$. Bold marks the best result.}
\label{tab:downstream}
\begin{tabular*}{\columnwidth}{@{\extracolsep{\fill}}lrrrr@{}}
\toprule
Reconstruction & Mean speed $\downarrow$ & LDJ $\downarrow$ & Closures $\downarrow$ & mAP $\uparrow$ \\
 & cm/s & & count & \% \\
\midrule
MS-TCN & 0.1729 & 5.911 & 0.0672 & 35.10 \\
PatchTST & 0.1204 & 5.174 & 0.0715 & 38.53 \\
Transformer regression & 0.1268 & 4.940 & 0.0694 & 41.57 \\
LiveMAE & 0.1163 & 5.445 & 0.0457 & 44.54 \\
\textbf{Proposed} & \textbf{0.0879} & \textbf{0.688} & \textbf{0.0414} & \textbf{54.44} \\
\bottomrule 
\end{tabular*}
\end{table}

\begin{figure}[t]
\centering
\includegraphics[width=\columnwidth]{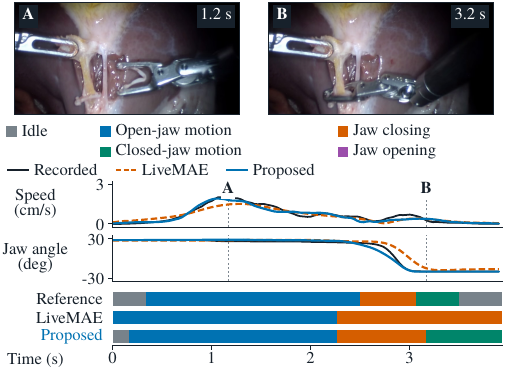}
\caption{Illustrative PSM1 episode in Open-H. Frames A and B show the open and closed grasper; dashed guides mark timestamps. Traces compare recorded, LiveMAE and proposed speed and jaw angle. Class strips compare recorded-state labels and MS-TCN predictions. Our reconstruction recovers closed-jaw motion omitted by LiveMAE. Episode F1 at 25\% temporal IoU is 88.9\% versus 57.1\% using the same classifier.} \vspace{-5mm} 
\label{fig:downstream}
\end{figure}

\textbf{Kinematic reconstruction comparison.}
We compare our method with four video reconstruction baselines using each
method's visual representation.
MS-TCN, PatchTST and a four-block Transformer encoder estimate instrument
position, orientation and jaw from global descriptors learned from frozen DINOv3
features using position supervision. MS-TCN uses two stages of temporal convolutional refinement; PatchTST encodes
temporal patches. Both use regression heads for the eight state coordinates. All three use constant-velocity RTS smoothing~\cite{RAUCH_1965}
with coefficients selected on validation data.
Our reconstruction network adds instrument landmarks, local appearance, depth, visual state
estimates and learned motion constraints.
LiveMAE jointly reconstructs images and recorded states with a
pretrained masked autoencoder and separate decoders~\cite{10.1007/978-3-031-43996-4_68};
validation state error selects its checkpoint.
Table~\ref{tab:main} reports the lowest errors for our method on all five
metrics. Path MAE is 0.3392\,cm for our approach versus 0.4510\,cm for LiveMAE, a 24.79\%
reduction. This metric compares traveled distance independently of position alignment.

\textbf{Temporal estimation with shared visual features.}
We compare six regressors using our visual representation, initial orientation
and jaw estimates, output parameterization and direct state loss.
Table~\ref{tab:ablation} includes validation-selected RTS postprocessing.
Our method achieves the lowest error on all five metrics. 

\textbf{Reconstruction and local temporal refinement.}
We evaluate weighted reconstruction and position and jaw refinement with training
ablations and component removal. Training ablations precede these refinements
and share inputs, initialization, loss terms and checkpoint selection.
Inference removals use the complete checkpoint without refitting. Position removal
disables both position heads and restores the position coefficient learned before refinement;
increment calibration and jaw projection remain active.
In Table~\ref{tab:ablation}, position refinement reduces path MAE from
0.3854 to 0.3392\,cm, an 11.98\% reduction. Removing reconstruction raises all five errors.

\textbf{Additional Open-H tasks and camera views.}
We evaluate additional tasks and views by fitting separate CAO and Hamlyn networks.
CAO uses the full visual representation; Hamlyn uses global appearance.
Direct state regression and its RTS-smoothed counterpart retain these inputs
and omit weighted reconstruction and position and jaw refinement.
Table~\ref{tab:extensions} reports cohort errors, and
Fig.~\ref{fig:trajectories} shows recorded frames, state traces and per-episode errors.
Our method achieves the lowest error for all five metrics in both cohorts.
On Hamlyn, path-length MAE is 1.2627\,cm versus 7.2639\,cm for direct state
regression with validation-selected RTS, an 82.62\% reduction.

\textbf{Evaluation on additional recordings.}
On 70 additional JHU episodes, our fixed main-benchmark network achieves the
lowest error on all five metrics (Table~\ref{tab:extensions}). Our approach achieves a Path MAE of
0.7751\,cm versus 0.9684\,cm for LiveMAE.

\textbf{Motion analysis from reconstructed states.}
Our reconstructions achieve the lowest indicator errors and highest
temporal mAP in Table~\ref{tab:downstream}. LDJ MAE is 0.688 versus 4.940
for Transformer regression using global DINOv3 features, an 86.07\% reduction.
Temporal mAP is 54.44\%, compared with 44.54\% for LiveMAE.
Fig.~\ref{fig:downstream} illustrates the reconstructed motion and segmentation. \vspace{-2mm} 

\section{Conclusion}
Our reconstruction network recovers instrument kinematics from monocular video using surgical visual features, separate displacement magnitude and direction prediction, and differentiable trajectory reconstruction. On the main Open-H benchmark, our method outperforms all baselines on all five reconstruction metrics. Path-length MAE is 0.3392\,cm, an 8.12\% reduction from 0.3692\,cm for Transformer regression using our visual representation and a 24.79\% reduction from 0.4510\,cm for LiveMAE. Log dimensionless jerk MAE is 0.688 versus 4.940 for Transformer regression using global DINOv3 features, an 86.07\% reduction. Motion segmentation achieves 54.44\% temporal mean average precision versus 44.54\% for LiveMAE.

\section*{Acknowledgements}

Generative AI (ChatGPT) was used for editing and grammar enhancement.

\bibliographystyle{researchIEEE}
\bibliography{refs}

\begin{thebibliography}{10}
\providecommand{\url}[1]{#1}
\csname url@rmstyle\endcsname
\providecommand{\newblock}{\relax}
\providecommand{\bibinfo}[2]{#2}
\providecommand\BIBentrySTDinterwordspacing{\spaceskip=0pt\relax}
\providecommand\BIBentryALTinterwordstretchfactor{4}
\providecommand\BIBentryALTinterwordspacing{\spaceskip=\fontdimen2\font plus
\BIBentryALTinterwordstretchfactor\fontdimen3\font minus \fontdimen4\font\relax}
\providecommand\BIBforeignlanguage[2]{{%
\expandafter\ifx\csname l@#1\endcsname\relax
\typeout{** WARNING: IEEEtran.bst: No hyphenation pattern has been}%
\typeout{** loaded for the language `#1'. Using the pattern for}%
\typeout{** the default language instead.}%
\else
\language=\csname l@#1\endcsname
\fi
#2}}

\bibitem{Ballo2026}
\BIBentryALTinterwordspacing
M.~Ballo, \emph{et~al.}, ``Semantic taxonomy-driven instrument classification streamlines kinematic analysis of objective performance indicators in robotic surgery,'' \emph{npj Digital Surgery}, vol.~1, no.~1, p.~6, 2026.
\BIBentrySTDinterwordspacing

\bibitem{Balasubramanian2015}
\BIBentryALTinterwordspacing
S.~Balasubramanian, \emph{et~al.}, ``On the analysis of movement smoothness,'' \emph{Journal of NeuroEngineering and Rehabilitation}, vol.~12, no.~1, p. 112, 2015.
\BIBentrySTDinterwordspacing

\bibitem{Chen_2025}
J.-T. Chen, \emph{et~al.}, ``SurgiPose: Estimating Surgical Tool Kinematics from Monocular Video for Surgical Robot Learning,'' in \emph{2025 IEEE/RSJ International Conference on Intelligent Robots and Systems (IROS)}, 2025, pp. 20912 through 20919.

\bibitem{10.1007/978-3-031-43996-4_68}
L.~Trinh, \emph{et~al.}, ``Self-supervised Sim-to-Real Kinematics Reconstruction for Video-Based Assessment of Intraoperative Suturing Skills,'' in \emph{Medical Image Computing and Computer Assisted Intervention, MICCAI 2023}, 2023, pp. 708 through 717.

\bibitem{consortium2026openhembodimentlargescaledatasetenabling}
\BIBentryALTinterwordspacing
Open-H-Embodiment Consortium, \emph{et~al.}, ``Open-H-Embodiment: A Large-Scale Dataset for Enabling Foundation Models in Medical Robotics,'' arXiv:2604.21017, 2026.
\BIBentrySTDinterwordspacing

\bibitem{Yue_2024}
\BIBentryALTinterwordspacing
W.~Yue, \emph{et~al.}, ``SurgicalSAM: Efficient Class Promptable Surgical Instrument Segmentation,'' \emph{Proceedings of the AAAI Conference on Artificial Intelligence}, vol.~38, no.~7, pp. 6890 through 6898, 2024.
\BIBentrySTDinterwordspacing

\bibitem{yue2024surgicalpartsamparttowholecollaborativeprompting}
\BIBentryALTinterwordspacing
W.~Yue, \emph{et~al.}, ``SurgicalPart-SAM: Part-to-Whole Collaborative Prompting for Surgical Instrument Segmentation,'' arXiv:2312.14481, 2024.
\BIBentrySTDinterwordspacing

\bibitem{liu2025surgicalsam2realtime}
\BIBentryALTinterwordspacing
H.~Liu, \emph{et~al.}, ``Surgical SAM 2: Real-time Segment Anything in Surgical Video by Efficient Frame Pruning,'' arXiv:2408.07931, 2025.
\BIBentrySTDinterwordspacing

\bibitem{dong2025segmentationbenchmarkingsam3}
\BIBentryALTinterwordspacing
W.~Dong, \emph{et~al.}, ``More than Segmentation: Benchmarking SAM 3 for Segmentation, 3D Perception, and Reconstruction in Robotic Surgery,'' arXiv:2512.07596, 2025.
\BIBentrySTDinterwordspacing

\bibitem{liu2026parameterefficientadaptationsam3promptdriven}
\BIBentryALTinterwordspacing
C.~Liu, \emph{et~al.}, ``Parameter-Efficient Adaptation of SAM3 for Prompt-Driven Surgical Concept Segmentation,'' arXiv:2607.23694, 2026.
\BIBentrySTDinterwordspacing

\bibitem{Du_2018}
\BIBentryALTinterwordspacing
X.~Du, \emph{et~al.}, ``Articulated Multi-Instrument 2-D Pose Estimation Using Fully Convolutional Networks,'' \emph{IEEE Transactions on Medical Imaging}, vol.~37, no.~5, pp. 1276 through 1287, 2018.
\BIBentrySTDinterwordspacing

\bibitem{Allan_2018}
\BIBentryALTinterwordspacing
M.~Allan, \emph{et~al.}, ``3-D Pose Estimation of Articulated Instruments in Robotic Minimally Invasive Surgery,'' \emph{IEEE Transactions on Medical Imaging}, vol.~37, no.~5, pp. 1204 through 1213, 2018.
\BIBentrySTDinterwordspacing

\bibitem{Xu_2025}
\BIBentryALTinterwordspacing
H.~Xu, \emph{et~al.}, ``SurgRIPE challenge: Benchmark of surgical robot instrument pose estimation,'' \emph{Medical Image Analysis}, vol. 105, p. 103674, 2025.
\BIBentrySTDinterwordspacing

\bibitem{Kalman_1960}
\BIBentryALTinterwordspacing
R.~E. Kalman, ``A New Approach to Linear Filtering and Prediction Problems,'' \emph{Journal of Basic Engineering}, vol.~82, no.~1, pp. 35 through 45, 1960.
\BIBentrySTDinterwordspacing

\bibitem{RAUCH_1965}
H.~E. Rauch, F.~Tung, and C.~T. Striebel, ``Maximum likelihood estimates of linear dynamic systems,'' \emph{AIAA journal}, vol.~3, no.~8, pp. 1445 through 1450, 1965.

\bibitem{Sarkka_2021}
\BIBentryALTinterwordspacing
S.~Sarkka and A.~F. Garcia-Fernandez, ``Temporal Parallelization of Bayesian Smoothers,'' \emph{IEEE Transactions on Automatic Control}, vol.~66, no.~1, pp. 299 through 306, 2021.
\BIBentrySTDinterwordspacing

\bibitem{NEURIPS2021_89fcd07f}
Z.~Teed and J.~Deng, ``DROID-SLAM: Deep Visual SLAM for Monocular, Stereo, and RGB-D Cameras,'' in \emph{Advances in Neural Information Processing Systems}, vol.~34, 2021, pp. 16558 through 16569.

\bibitem{NEURIPS2022_18596929}
L.~Pineda, \emph{et~al.}, ``Theseus: A Library for Differentiable Nonlinear Optimization,'' in \emph{Advances in Neural Information Processing Systems}, vol.~35, 2022, pp. 3801 through 3818.

\bibitem{ICLR2026_e0982cbc}
N.~Carion, \emph{et~al.}, ``SAM 3: Segment Anything with Concepts,'' in \emph{International Conference on Learning Representations}, 2026, pp. 138846 through 138923.

\bibitem{Alabi2025}
\BIBentryALTinterwordspacing
O.~Alabi, \emph{et~al.}, ``CholecInstanceSeg: A Tool Instance Segmentation Dataset for Laparoscopic Surgery,'' \emph{Scientific Data}, vol.~12, no.~1, p. 825, 2025.
\BIBentrySTDinterwordspacing

\bibitem{hong2020cholecseg8ksemanticsegmentationdataset}
\BIBentryALTinterwordspacing
W.~Y. Hong, \emph{et~al.}, ``CholecSeg8k: A Semantic Segmentation Dataset for Laparoscopic Cholecystectomy Based on Cholec80,'' arXiv:2012.12453, 2020.
\BIBentrySTDinterwordspacing

\bibitem{Carstens2023}
\BIBentryALTinterwordspacing
M.~Carstens, \emph{et~al.}, ``The Dresden Surgical Anatomy Dataset for Abdominal Organ Segmentation in Surgical Data Science,'' \emph{Scientific Data}, vol.~10, no.~1, p.~3, 2023.
\BIBentrySTDinterwordspacing

\bibitem{murali2024endoscapesdatasetsurgicalscene}
\BIBentryALTinterwordspacing
A.~Murali, \emph{et~al.}, ``The Endoscapes Dataset for Surgical Scene Segmentation, Object Detection, and Critical View of Safety Assessment: Official Splits and Benchmark,'' arXiv:2312.12429, 2024.
\BIBentrySTDinterwordspacing

\bibitem{Oh_2025}
\BIBentryALTinterwordspacing
K.-H. Oh, \emph{et~al.}, ``Expanded Comprehensive Robotic Cholecystectomy Dataset,'' \emph{Journal of Medical Robotics Research}, vol.~10, no. 03n04, 2025.
\BIBentrySTDinterwordspacing

\bibitem{siméoni2025dinov3}
\BIBentryALTinterwordspacing
O.~Siméoni, \emph{et~al.}, ``DINOv3,'' arXiv:2508.10104, 2025.
\BIBentrySTDinterwordspacing

\bibitem{10.1007/978-3-032-08009-7_12}
N.~Toussaint, \emph{et~al.}, ``Zero-Shot Monocular Metric Depth for Endoscopic Images,'' in \emph{Data Engineering in Medical Imaging}, 2026, pp. 115 through 124.

\bibitem{Zhang_2010}
Y.~Zhang, J.~Cohen, and J.~D. Owens, ``Fast tridiagonal solvers on the GPU,'' in \emph{Proceedings of the 15th ACM SIGPLAN Symposium on Principles and Practice of Parallel Programming}, 2010, pp. 127 through 136.

\bibitem{Farha_2019_CVPR}
Y.~A. Farha and J.~Gall, ``MS-TCN: Multi-Stage Temporal Convolutional Network for Action Segmentation,'' in \emph{Proceedings of the IEEE/CVF Conference on Computer Vision and Pattern Recognition (CVPR)}, 2019.

\bibitem{Yuqietal-2023-PatchTST}
Y.~Nie, \emph{et~al.}, ``A Time Series is Worth 64 Words: Long-term Forecasting with Transformers,'' in \emph{International Conference on Learning Representations}, 2023.

\bibitem{NIPS2017_3f5ee243}
A.~Vaswani, \emph{et~al.}, ``Attention is All you Need,'' in \emph{Advances in Neural Information Processing Systems}, vol.~30, 2017.

\bibitem{cho-etal-2014-learning}
K.~Cho, \emph{et~al.}, ``Learning Phrase Representations using {RNN} Encoder{-}Decoder for Statistical Machine Translation,'' in \emph{Proceedings of the 2014 Conference on Empirical Methods in Natural Language Processing ({EMNLP})}, 2014, pp. 1724 through 1734.

\bibitem{Hochreiter_1997}
\BIBentryALTinterwordspacing
S.~Hochreiter and J.~Schmidhuber, ``Long Short-Term Memory,'' \emph{Neural Computation}, vol.~9, no.~8, pp. 1735 through 1780, 1997.
\BIBentrySTDinterwordspacing

\bibitem{bai2018empiricalevaluationgenericconvolutional}
\BIBentryALTinterwordspacing
S.~Bai, J.~Z. Kolter, and V.~Koltun, ``An Empirical Evaluation of Generic Convolutional and Recurrent Networks for Sequence Modeling,'' arXiv:1803.01271, 2018.
\BIBentrySTDinterwordspacing

\bibitem{Kabsch_1976}
\BIBentryALTinterwordspacing
W.~Kabsch, ``A solution for the best rotation to relate two sets of vectors,'' \emph{Acta Crystallographica Section A}, vol.~32, no.~5, pp. 922 through 923, 1976.
\BIBentrySTDinterwordspacing

\end{thebibliography}
\end{document}